\documentclass[
twocolumn,
]{ceurart}

\usepackage{listings}
\usepackage{hyperref}

\newcommand{\set}[1]{\mathcal{#1}}
\providecommand{\sE}{\ensuremath{\set{E}}}
\providecommand{\sG}{\ensuremath{\set{G}}}
\providecommand{\sV}{\ensuremath{\set{V}}}
\providecommand{\sW}{\ensuremath{\set{W}}}
\providecommand{\sX}{\ensuremath{\set{X}}}

\newcommand{\mat}[1]{\mathbf{#1}}
\providecommand{\mA}{\ensuremath{\mat{A}}}
\providecommand{\mB}{\ensuremath{\mat{B}}}
\providecommand{\mC}{\ensuremath{\mat{C}}}
\providecommand{\mD}{\ensuremath{\mat{D}}}
\providecommand{\mX}{\ensuremath{\mat{X}}}
\providecommand{\mW}{\ensuremath{\mat{W}}}
\providecommand{\mZ}{\ensuremath{\mat{Z}}}
\providecommand{\mbA}{\ensuremath{\overline{\mat{A}}}}
\providecommand{\mbB}{\ensuremath{\overline{\mat{B}}}}
\providecommand{\mbC}{\ensuremath{\overline{\mat{C}}}}
\providecommand{\mbK}{\ensuremath{\overline{\mat{K}}}}

\renewcommand{\vec}[1]{{\bf{#1}}}
\providecommand{\vh}{\ensuremath{\vec{h}}}
\providecommand{\vx}{\ensuremath{\vec{x}}}
\providecommand{\vy}{\ensuremath{\vec{y}}}
\providecommand{\vw}{\ensuremath{\vec{w}}}

\newcolumntype{C}[1]{>{\centering\arraybackslash}p{#1}}

\newcommand{\ours}[0]{\textup{SpoT-Mamba}\xspace}

\newcommand{\pems}[0]{\textsl{PEMS04}\xspace}

\newcommand{\hi}[0]{\textup{HI}\xspace}
\newcommand{\gwnet}[0]{\textup{GWNet}\xspace}
\newcommand{\dcrnn}[0]{\textup{DCRNN}\xspace}
\newcommand{\agcrn}[0]{\textup{AGCRN}\xspace}
\newcommand{\stgcn}[0]{\textup{STGCN}\xspace}
\newcommand{\gts}[0]{\textup{GTS}\xspace}
\newcommand{\mtgnn}[0]{\textup{MTGNN}\xspace}
\newcommand{\stnorm}[0]{\textup{STNorm}\xspace}
\newcommand{\gman}[0]{\textup{GMAN}\xspace}
\newcommand{\pdformer}[0]{\textup{PDformer}\xspace}
\newcommand{\stid}[0]{\textup{STID}\xspace}
\newcommand{\stae}[0]{\textup{STAEformer}\xspace}

\begin{document}

\copyrightyear{2024}
\copyrightclause{Copyright for this paper by its authors.
  Use permitted under Creative Commons License Attribution 4.0
  International (CC BY 4.0).}

\conference{STRL'24: Third International Workshop on Spatio-Temporal Reasoning and Learning, 5 August 2024, Jeju, South Korea}

\title{SpoT-Mamba: Learning Long-Range Dependency on Spatio-Temporal Graphs with Selective State Spaces}

\author[1]{Jinhyeok Choi}[
email=cjh0507@kaist.ac.kr
]
\address[1]{School of Computing, KAIST, Daejeon, Republic of Korea}

\author[1]{Heehyeon Kim}[
email=heehyeon@kaist.ac.kr
]

\author[1]{Minhyeong An}[
email=amh0360@kaist.ac.kr
]

\author[1]{Joyce Jiyoung Whang}[
email=jjwhang@kaist.ac.kr
]
\cormark[1]

\cortext[1]{Corresponding author.}

\begin{abstract}
    Spatio-temporal graph (STG) forecasting is a critical task with extensive applications in the real world, including traffic and weather forecasting. Although several recent methods have been proposed to model complex dynamics in STGs, addressing long-range spatio-temporal dependencies remains a significant challenge, leading to limited performance gains. Inspired by a recently proposed state space model named Mamba, which has shown remarkable capability of capturing long-range dependency, we propose a new STG forecasting framework named \ours. \ours generates node embeddings by scanning various node-specific walk sequences. Based on the node embeddings, it conducts temporal scans to capture long-range spatio-temporal dependencies. Experimental results on the real-world traffic forecasting dataset demonstrate the effectiveness of \ours.    
\end{abstract}



\maketitle

\section{Introduction}
The predictive learning methods on time series play a crucial role in diverse applications, such as traffic and weather forecasting. The intricate relationships and dynamic nature of time series are often represented as graphs, specifically spatio-temporal graphs (STGs), where node attributes evolve over time. Recently, spatio-temporal graph neural networks (STGNNs) have emerged as a powerful tool for capturing both spatial and temporal dependencies in STGs~\cite{dgcnn, stem, sftgnn}. Many of those methods employ graph neural networks (GNNs) to exploit spatial dependencies inherent in the graph structures, integrating them with recurrent units or convolutions to capture temporal dependencies~\cite{stem, zigzag, hgcn, stag, cgc, tgcn, gwnet}. These approaches have facilitated the capturing of spatio-temporal dependencies within STGs. Despite their remarkable performance in predictive learning tasks, they often face challenges in handling long-range temporal dependencies among different time steps~\cite{info, fed}.

STGs often exhibit repetitive patterns over both short and long periods, which is critical for precise predictions~\cite{pyra, lstnet}. Therefore, several methods have adopted self-attention mechanisms of transformer layers~\cite{att} rather than recurrent units to enhance their capability in exploiting global temporal information~\cite{info, fed, astgcn}. However, the significant computational overhead and complexity of attention mechanisms are being highlighted as major concerns~\cite{info, pyra, hi, agnostic}.

Meanwhile, structured state space sequence (S4) models have emerged as a promising approach for sequence modeling with linear scaling in sequence length~\cite{s4}. Those models take the advantages of recurrent neural networks and convolutional neural networks, enabling them to handle long-range dependencies without relying on attention. However, due to their inability to select information depending on input data, they have shown limited performance.

A recent study has introduced a new S4 model overcoming the issue, named Mamba, which introduces a selection mechanism to filter information in an input-dependent manner~\cite{mamba}. Mamba has demonstrated notable performance over transformers across various types of sequence data, including language, audio, and genomics. In addition, there have been several studies towards replacing the transformer with Mamba in graph transformer frameworks~\cite{gmamba1, gmamba2, stgmamba}.

In this paper, our focus lies on the predictive learning task on STGs, specifically STG forecasting. For STG forecasting, it is vital to capture the evolving behavior of individual nodes over time and how these changes propagate throughout the entire graph. Furthermore, leveraging these dynamics over long spatial and temporal ranges plays a crucial role in dealing with the intricate spatio-temporal correlations in STGs~\cite{dgcnn, sftgnn, stgode}. Building upon these insights and recent advances, we introduce \ours, a new \textbf{Sp}ati\textbf{o}-\textbf{T}emporal graph forecasting framework with a \textbf{Mamba}-based sequence modeling architecture. With Mamba blocks, \ours extracts structural information of each node by scanning multi-way walk sequences and effectively captures long-range temporal dependencies with temporal scans. Experiments on the real-world dataset demonstrate that \ours achieves promising performance in STG forecasting. The official implementations of \ours are available at \href{https://github.com/bdi-lab/SpoT-Mamba}{https://github.com/bdi-lab/SpoT-Mamba}.

\section{Preliminaries}
\paragraph{\normalfont{\textbf{State Space Model (SSM)}}}
The state space model (SSM) assumes that dynamic systems can be represented by their states at time step $t$~\cite{s4}. SSM defines the evolution of a dynamic system's state with two equations: $\vh'(t) = \mA \vh(t)+ \mB x(t)$ and $y(t) = \mC \vh(t)+ \mD x(t)$, where $\vh(t) \in \mathbb{R}^D$ denotes the latent state, $x(t) \in \mathbb{R}$ represents the input signal, $y(t)\in \mathbb{R}$ denotes the output signal, and $\mA \in \mathbb{R}^{D \times D}, \mB \in \mathbb{R}^{D \times 1}, \mC \in \mathbb{R}^{D \times D}$, and $\mD \in \mathbb{R}$ are learnable parameters. SSM learns how to transform the input signal $x(t)$ into the latent state $\vh(t)$, which is used to model the system dynamics and predict its output $y(t)$.

\paragraph{\normalfont{\textbf{Discretized SSM}}}
To adapt SSM for discrete input sequences instead of continuous signals, discretization is applied with a step size $\Delta$. The discretized SSM is defined in a recurrent form: $\vh_{t}=\mbA \vh_{t-1} + \mbB x_{t}$ and $y_{t}=\mbC \vh_{t}$, where $\mbA$ and $\mbB$ are approximated learnable parameters using a bilinear method with a step size $\Delta$~\cite{s4}. The term $\mD$ is omitted from the equations as it can be considered as a skip connection.  This formulation allows for capturing temporal dependencies efficiently, resulting in similar computations to those in recurrent neural networks.


Meanwhile, due to its linear time-invariant (LTI) property, SSM can be reformulated as discrete convolution: $\vy=\mbK \ast \vx$ and $\mbK \in \mathbb{R}^{L}=(\mbC\mbB, \mbC\mbA\mbB, \dots, \mbC\mbA^{L-1}\mbB)$, where $\vx \in \mathbb{R}^{L}$ denotes the input sequence, $\vy \in \mathbb{R}^{L}$ denotes the output sequence, $\ast$ indicates the convolution operation, and $L$ is the sequence length. This representation facilitates parallel training for SSM, thereby enhancing training efficiency. 

The recurrent and convolutional representations of SSM for sequence modeling enable parallel training and linear scaling in sequence length. To further enhance the computational complexity of SSM, the structured state space sequence (S4) models have been proposed~\cite{s4}. S4 models address the fundamental bottleneck of SSM, which involves repeated matrix multiplications, by employing a low-rank correction to stably diagonalize the transition matrix $\mA$.



\paragraph{\normalfont{\textbf{Mamba}}} 
S4 models have demonstrated remarkable performance in handling long-range dependencies in continuous signal data, such as audio and time series. However, S4 models struggle with effectively handling discrete and information-dense data such as text~\cite{mamba}. This limitation arises from the LTI property inherent in the convolutional form of SSMs. While the LTI property enables linear time sequence modeling for S4 models, it requires that the learnable matrices $\mA$, $\mB$, and $\mC$, as well as the step size $\Delta$, remain unchanged across all time steps. Consequently, S4 models cannot selectively recall previous tokens or combine the current token, treating each token in the input sequence uniformly. In contrast, Transformers dynamically adjust attention scores based on the input sequence, allowing them to effectively focus on different parts of the sequence~\cite{att}.

To address both the lack of selectivity in S4 models and the efficiency bottleneck in sequence modeling, a recent study introduced a new S4 model called Mamba, which removes the LTI constraints~\cite{mamba}. Mamba incorporates a selection mechanism that allows its learnable parameters to dynamically interact with the input sequence. This mechanism is achieved by modifying the learnable parameters $\mB$ and $\mC$, as well as the step size $\Delta$, to functions of the input sequence. Therefore, Mamba can selectively recall or ignore information in an input-dependent manner, while maintaining linear scalability in sequence length.

Inspired by the recent advancements in Mamba, we propose a Mamba-based sequence modeling architecture for predictive learning tasks on STGs. Our approach employs a selective mechanism to handle the dynamical changes in STGs, capturing long-range spatio-temporal dependencies. In addition, this allows for addressing  computational inefficiencies in transformer-based STGNNs~\cite{info, fed, astgcn}.


\begin{figure*}[t]
\includegraphics[width=2.1\columnwidth]{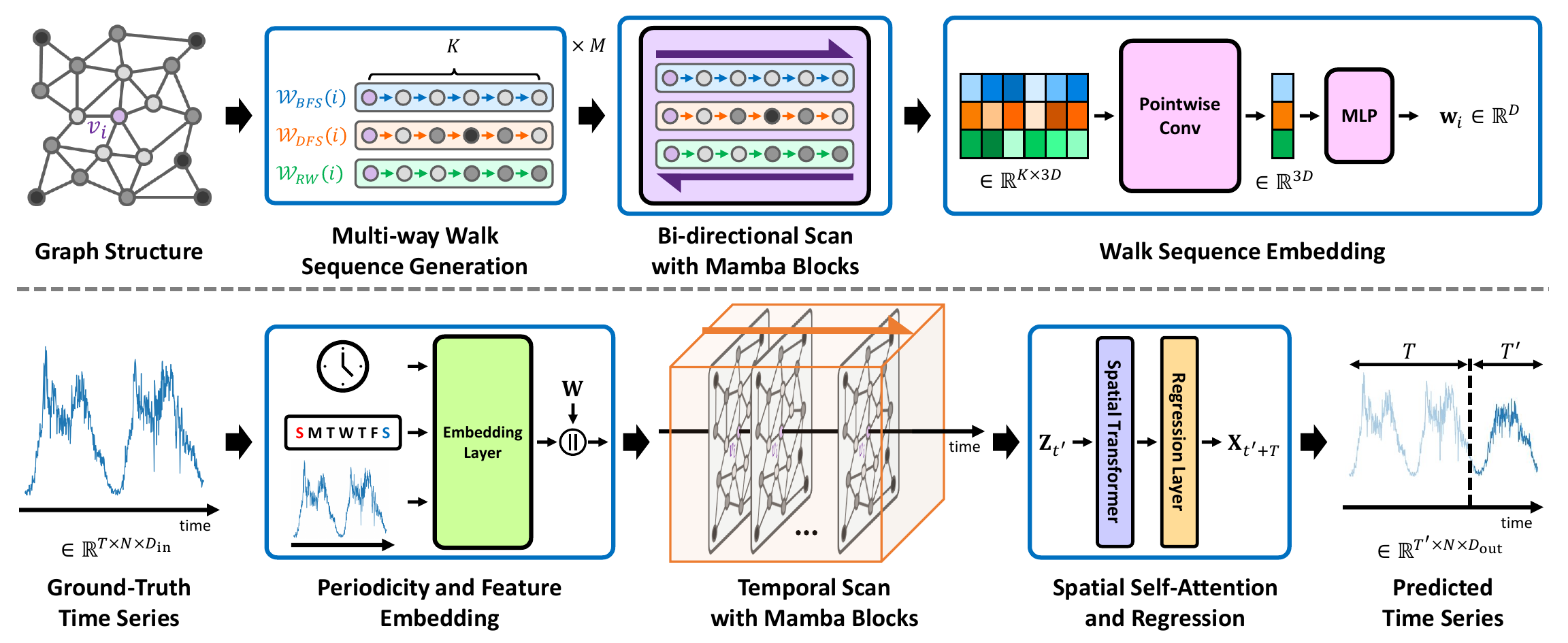}
\caption{Overview of \ours. The first row represents the node-specific walk sequence embedding. The second row represents the overall procedure of STG forecasting. $\mW \in \mathbb{R}^{N \times D}$ denotes the node embeddings for all nodes in the graph and $\mZ_{t'} \in \mathbb{R}^{N \times 4D}$ denotes one of the outcomes from the temporal scan, corresponding to the time step $t'$.}
\label{fig:ours}
\end{figure*}

\paragraph{\normalfont{\textbf{Spatio-Temporal Graph Forecasting}}}
A spatio-temporal graph (STG) is defined as $\sG = (\sV, \sE, \sX)$, where $\sV$ is a set of $N$ nodes, $\sE \subset \sV \times \sV$ is a set of edges, $\sX = [\mX_1, \dots, \mX_{\tau}]$ is a sequence of observed data for all nodes at each historical time step, and $\tau$ is a length of the sequence. Here, $\mX_t \in \mathbb{R}^{N \times D_{\text{in}}}$ denotes the observed data at time step $t$, where $D_{\text{in}}$ denotes the dimension of the input node attributes. STG forecasting aims to predict future observations for $T'$ time steps, given historical observations for the previous $T$ time steps. This is formulated as $[\mX_{t-T+1}, \dots, \mX_t] \xrightarrow{f(\cdot)} [\mX_{t+1}, \dots, \mX_{t+T'}]$, where $f(\cdot)$ represents the STG forecasting model.

\section{Spatio-Temporal Graph Forecasting with \ours} 
We propose \ours (Figure~\ref{fig:ours}), which captures spatial dependencies from node-specific walk sequences and learns temporal dependencies across time steps leveraging Mamba-based sequence modeling. By utilizing the selection mechanisms of Mamba blocks, \ours can selectively propagate or forget information in an input-dependent manner on both temporal and spatial domain.

\paragraph{\normalfont{\textbf{Multi-way Walk Sequence}}}
In STGs, the temporal sequences for nodes are naturally defined by the time-series
data. On the other hand, since the topological structure does not have a specific order, a tailored method is required to define the spatial sequences of nodes in graphs. Hence, we employ three well-known walk algorithms: depth-first search (DFS), breadth-first search (BFS), and random walks (RW), to extract diverse local and global structural information from each node’s neighborhood. The walk sequences of length $K$ for node $v_i$ using these walk algorithms are defined as $\sW_{BFS}(i)$, $\sW_{DFS}(i)$, and $\sW_{RW}(i)$, respectively. These node-specific walk sequences are extracted $M$ times to exploit more comprehensive structural information.


\paragraph{\normalfont{\textbf{Walk Sequence Embedding}}}
\ours generates embeddings for node-specific walk sequences by scanning each sequence. Here, \ours performs bi-directional scans through Mamba blocks, which makes the model robust to permutations and captures the long-range spatial dependency of the sequence more effectively~\cite{gmamba2}. Then, \ours aggregates the representations of node-specific walk sequences with pointwise convolution. This allows for incorporating representations of neighboring nodes to generate walk sequence embedding for each type of walk sequence.

Subsequently, \ours integrates the walk sequence embeddings into a node embedding $\vw_{i} \in \mathbb{R}^{D}$ using Multi-Layer Perceptron (MLP), where $i$ represents the node index and $D$ denotes the embedding dimension. Rather than simply stacking GNN layers, we employ Mamba-based sequence modeling to generate node embeddings from diverse types of walk sequences. Therefore, our approach effectively captures local and long-range dependencies within the graph by scanning the neighborhood structure of each node.


\paragraph{\normalfont{\textbf{Temporal Scan with Mamba Blocks}}}
We adopt the learnable day-of-week and timestamps-of-day embeddings to capture the repetitive patterns over both short and long periods in STGs~\cite{stae}. \ours concatenates these embeddings with the node embedding for each time step to effectively model temporal dynamics from historical observations. Then, it performs selective recurrent scans with Mamba blocks across the sequence of embeddings along the time axis, observing changes in temporal dynamics over time. This process helps identify critical portions of the sequence for forecasting and captures periodic patterns in both short- and long-term intervals. Consequently, the results effectively encompass spatio-temporal dependencies, thereby enriching the predictive capabilities.


\paragraph{\normalfont{\textbf{STG forecasting of \ours}}}
Finally, \ours enhances the representations of nodes scanned along the temporal axis by incorporating global information from the entire graph at each time step through transformer layers. Then, MLP is applied to forecast the attributes of each node for future time steps. To accurately predict the temporal trajectory while ensuring robustness to outliers that deviate significantly from the expected trajectory, we train \ours utilizing the Huber loss, which is less sensitive to outliers while maintaining the smoothness of the squared error for small errors~\cite{stae}.

\section{Experiments}
We compare the performance of \ours with state-of-the-art baselines. Additionally, we conduct ablation studies to further demonstrate the effectiveness of \ours.

\subsection{Dataset and Experimental Setup}

\paragraph{\normalfont{\textbf{Dataset}}}
We evaluate \ours on \pems~\cite{stsgcn}, a real-world traffic flow forecasting benchmark, following the experimental setup in~\cite{stae}. \pems contains highway traffic flow data collected from the California Department of Transportation's Performance Measurement System (PEMS). The nodes represent the sensors, and edges are created when two sensors are on the same road. Traffic data in \pems is collected every 5 minutes. We set the input and prediction intervals to 1 hour, corresponding to $T = T' = 12$. The statistic of \pems is shown in Table~\ref{tab:dataset}. In the experiments, \pems is divided into training, validation, and test sets in a 6:2:2 ratio in temporal order.

\paragraph{\normalfont{\textbf{Baselines}}}
We compare \ours with several baselines using various methods, including GNNs and Transformers. For STGNNs, we consider \gwnet~\cite{gwnet}, \dcrnn~\cite{dcrnn}, \agcrn~\cite{agcrn}, \gts~\cite{gts}, and \mtgnn~\cite{mtgnn}. For attention-based methods, we include \stae~\cite{stae}, \gman~\cite{gman}, and \pdformer~\cite{pd}. Additionally, other methods such as \hi~\cite{hi}, \stnorm~\cite{stnorm}, and \stid~\cite{stid} are also considered.

\paragraph{\normalfont{\textbf{Evaluation Metrics}}}
We use three standard metrics for traffic flow prediction: Mean Absolute Error (MAE), Root Mean Squared Error (RMSE), and Mean Absolute Percentage Error (MAPE). MAE is the unweighted mean of the absolute differences between predictions and ground truth values. RMSE is calculated as the square root of the average of the squared differences. MAPE is similar to MAE but normalizes each error by the corresponding ground truth value and expresses it as a percentage. For all three metrics, lower values indicate better performance. Additionally, we rank all the methods used in our experiments across these evaluation metrics and compute their average ranks.

\paragraph{\normalfont{\textbf{Implementation Details}}}
\ours is implemented using the Deep Graph Library~\cite{dgl} and PyTorch~\cite{pytorch}. For the transformer, we utilize the off-the-shelf transformer encoder available in PyTorch, and for Mamba, we employ the official implementation~\cite{mamba} and apply pre-normalization. We train \ours for 300 epochs using the Adam optimizer~\cite{adam}, with early stopping if there is no improvement over 20 epochs. Additionally, we apply the learning rate decay, reducing the learning rate at the 20th, 40th, and 60th epochs. To determine the optimal hyperparameters for \ours, we conduct a grid search. The grid search covers $M \in \{2, 4\}$, learning rates of $\{0.001, 0.0005\}$), weight decays of $\{0.001, 0.0001\}$, and learning rate decay rates of $\{0.1, 0.5\}$. We fix the feed-forward dimension to 256, $D = 32$, $K = 20$, dropout probability to 0.1, batch size to 32, and the number of layers for Mamba and the transformer to 3. All experiments are conducted using GeForce RTX 3090 24GB.

\begin{table}[t]
\renewcommand{\arraystretch}{1.3}
\setlength{\tabcolsep}{0.38em}
\caption{Statistic of \pems dataset.}
\label{tab:dataset}
\begin{tabular}{ c|c|c|c|c }
\Xhline{2\arrayrulewidth}
    $\lvert \mathcal{V} \rvert$ & $\lvert \mathcal{E} \rvert$ & \#Time Steps & Time Interval & Time Range  \\
\Xhline{\arrayrulewidth}
    307 & 338 & 16,992 & 5 min. & 01/2018 - 02/2018 \\ 
\Xhline{2\arrayrulewidth}
\end{tabular}
\end{table}

\begin{table}[t]
\small
\renewcommand{\arraystretch}{1.10}
\setlength{\tabcolsep}{0.28em}
\caption{Traffic forecasting performance on \pems. `Avg.' denotes the average rank across the three evaluation metrics.}
\label{tab:result}

\begin{tabular}{ c|cc|cc|cc|C{2.2em} }
\Xhline{2\arrayrulewidth}
           & MAE & Rank & RMSE & Rank & MAPE & Rank & Avg. \\
\Xhline{\arrayrulewidth}
\hi        & 42.35 & 13 & 61.66 & 13 & 29.92 & 13 & 13   \\
\gwnet      & 18.53 & 5  & \textbf{29.92} & \textbf{1}  & 12.89 & 6  & 4    \\
\dcrnn     & 19.63 & 11 & 31.26 & 8  & 13.59 & 11 & 10   \\
\agcrn     & 19.38 & 9  & 31.25 & 7  & 13.40 & 9  & 8.33 \\
\stgcn     & 19.57 & 10 & 31.38 & 9  & 13.44 & 10 & 9.67 \\
\gts       & 20.96 & 12 & 32.95 & 12 & 14.66 & 12 & 12   \\
\mtgnn     & 19.17 & 8  & 31.70 & 11 & 13.37 & 8  & 9    \\
\stnorm    & 18.96 & 6  & 30.98 & 6  & 12.69 & 5  & 5.67 \\
\gman      & 19.14 & 7  & 31.60 & 10 & 13.19 & 7  & 8    \\
\pdformer  & 18.36 & 3  & 30.03 & 3  & 12.00 & 3  & 3    \\
\stid      & 18.38 & 4  & \underline{29.95} & \underline{2}  & 12.04 & 4  & 3.33 \\
\stae      & \textbf{18.22} & \textbf{1}  & 30.18 & 5  & \underline{11.98} & \underline{2} & 2.67 \\
\Xhline{\arrayrulewidth}
\ours      & \underline{18.31} & \underline{2}  & 30.11 & 4  & \textbf{11.86} & \textbf{1}  & \textbf{2.33} \\
\Xhline{2\arrayrulewidth}
\end{tabular}

\end{table}

\subsection{Traffic Forecasting Performance}
Table~\ref{tab:result} shows the traffic forecasting performance of the baselines and \ours on \pems in terms of MAE, RMSE, and MAPE. Note that we report the baseline results from \cite{stae} since we strictly followed the experimental settings described in \cite{stae}.  For each evaluation metric, the best results are boldfaced, and the second-best results are underlined. It is observed that \ours consistently achieves high rankings across all metrics: MAE, RMSE, and MAPE, recording the highest average rank among all methods. This suggests the effectiveness of Mamba's selective recurrent scan in modeling spatio-temporal dependency. Compared to other metrics, \ours demonstrates its best performance in MAPE, achieving the highest ranking. Among the baselines, \stae~\cite{stae} shows the most comparable performance to \ours.

\begin{figure*}[t]
\centering
\includegraphics[width=\columnwidth]{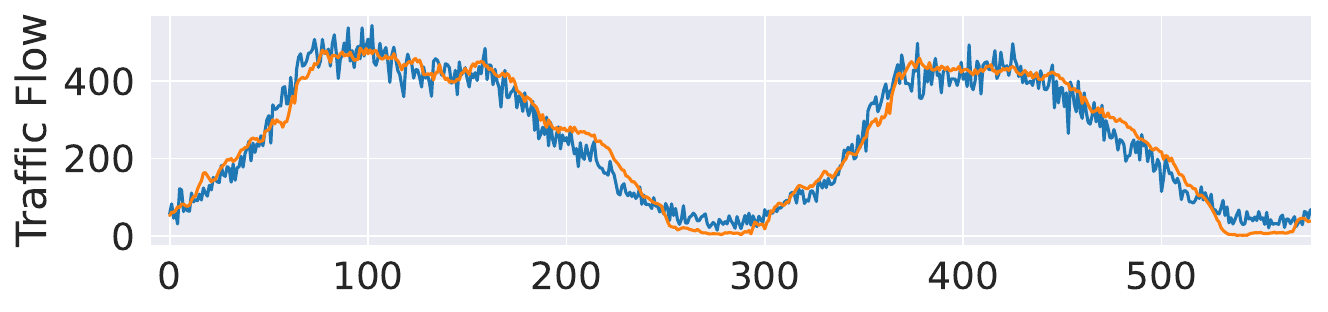}
\includegraphics[width=\columnwidth]{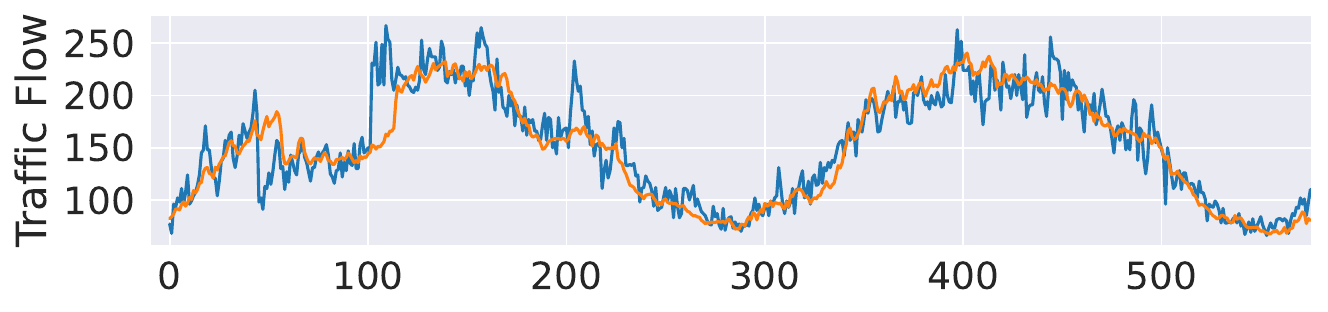}
\includegraphics[width=\columnwidth]{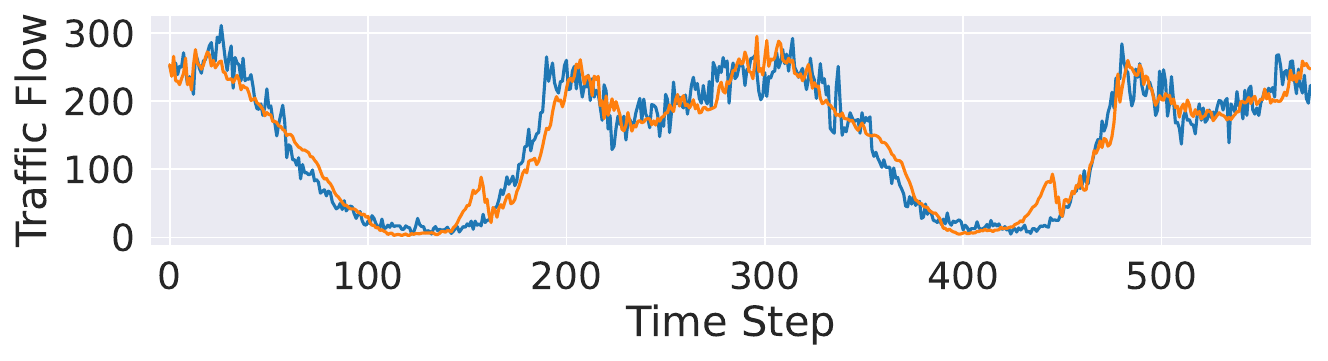}
\includegraphics[width=\columnwidth]{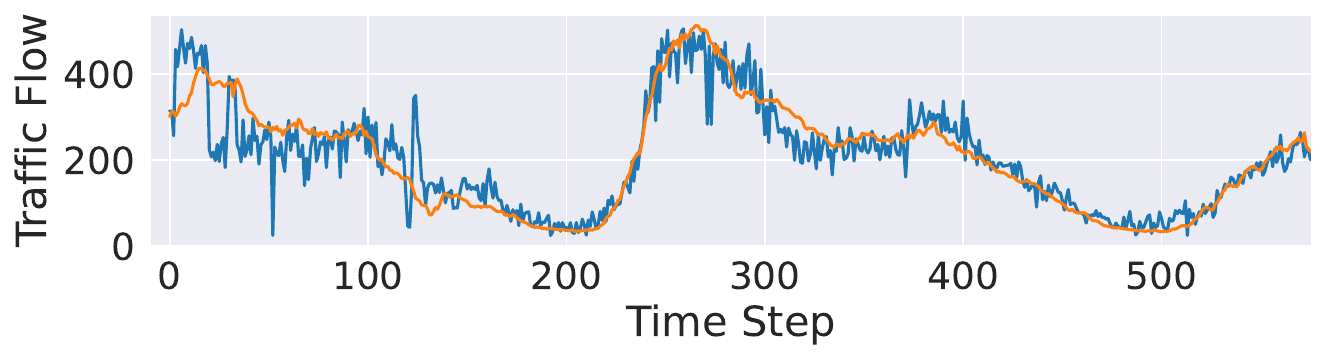}
\caption{Traffic flow forecasting results for four randomly selected nodes in \pems. The blue line represents the ground truth, and the orange line denotes the predictions made by \ours.}
\label{fig:pred}
\end{figure*}

\subsection{Qualitative Analysis \& Ablation Studies}
In Figure~\ref{fig:pred}, we visualize the predictions of \ours and the ground-truth time series on \pems. For this visualization, predictions are made for four randomly selected nodes, starting from an arbitrary time step within a test split, covering 576 consecutive time steps (equivalent to two days). Since we set $T=T'=12$, multiple predictions are concatenated to represent the duration. When multiple predictions exist at a single time step, we average them. It is observed that the predicted time series closely aligns with the ground-truth data.

Additionally, we conduct the ablation studies of \ours on \pems. We replace the Mamba blocks used for the walk sequence embedding (indicated as Walk Scan) and for scanning along the time axis (indicated as Temporal Scan) with transformer encoders. Note that when replacing the Mamba blocks for the Temporal Scan with a transformer encoder, we reduce the batch size from 32 to 8 due to Out-of-Memory issues. Results are shown in Table~\ref{tab:ablation}. We observed differences in performance depending on which type of scan module is replaced. Specifically, when the Walk Scan is conducted by a transformer encoder, the overall performance of \ours decreases (first and second rows). On the other hand, replacing only the Mamba blocks for the Temporal Scan with a transformer encoder shows negligible performance differences (third row). 

This disparity can be attributed to the inherent differences between Mamba and Transformer, along with the application of learnable embeddings that impose biases on the sequence. Mamba blocks scan inputs recurrently, inherently considering the sequence order. In contrast, the transformer encoder does not recognize input sequence order by itself. Furthermore, while \ours utilizes learnable embeddings for temporal sequences, i.e., day-of-week and timestamps-of-day, it does not apply such embeddings for walk sequences. As a result, the Transformer encoder struggles to perceive the order in walk sequences, despite performing well with temporal sequences.






\begin{table}[t]
\renewcommand{\arraystretch}{1.15}
\setlength{\tabcolsep}{0.38em}
\caption{Ablation studies of \ours on \pems, varying the types of the walk scan and temporal scan modules.}
\label{tab:ablation}
\begin{tabular}{ cc|C{2.7em}|C{2.7em}|C{2.7em} }
\Xhline{2\arrayrulewidth}
    Walk Scan & Temporal Scan & MAE & RMSE & MAPE  \\
\Xhline{\arrayrulewidth}
    Transformer & Transformer & 18.41 &  30.32  &  12.12 \\ 
    Transformer & Mamba & 18.69 & 30.17 & 12.28 \\ 
    Mamba & Transformer & 18.29 & 30.06 & 11.93 \\ 
\Xhline{\arrayrulewidth}
    Mamba & Mamba & 18.31 & 30.11 & 11.86 \\ 
\Xhline{2\arrayrulewidth}
\end{tabular}
\end{table}

\section{Conclusion and Future Work}
In this paper, we explore STG forecasting and introduce a new Mamba-based STG forecasting model, \ours. \ours utilizes Mamba blocks to scan multi-way walk sequences and temporal sequences. This approach allows the model to effectively capture the long-range spatio-temporal dependencies in STG, enhancing forecasting accuracy on complex graph structures. \ours shows promising results on the real-world traffic forecasting benchmark \pems. For future work, we will extend \ours to handle graphs with complex relations~\cite{drag, hynt} and evolving graphs~\cite{dgraph, ingram}. 



\begin{acknowledgments}
This research was supported by an NRF grant funded by  MSIT 2022R1A2C4001594 (Extendable Graph Representation Learning) and an IITP grant funded by MSIT 2022-0-00369 (Development of AI Technology to support Expert Decision-making that can Explain the Reasons/Grounds for  Judgment Results based on Expert Knowledge).
\end{acknowledgments}

\bibliography{ref_strl2024}

\begin{thebibliography}{40}
\expandafter\ifx\csname natexlab\endcsname\relax\def\natexlab#1{#1}\fi
\providecommand{\url}[1]{\texttt{#1}}
\providecommand{\href}[2]{#2}
\providecommand{\path}[1]{#1}
\providecommand{\DOIprefix}{doi:}
\providecommand{\ArXivprefix}{arXiv:}
\providecommand{\URLprefix}{URL: }
\providecommand{\Pubmedprefix}{pmid:}
\providecommand{\doi}[1]{\href{http://dx.doi.org/#1}{\path{#1}}}
\providecommand{\Pubmed}[1]{\href{pmid:#1}{\path{#1}}}
\providecommand{\bibinfo}[2]{#2}
\ifx\xfnm\relax \def\xfnm[#1]{\unskip,\space#1}\fi
\bibitem[{Diao et~al.(2019)Diao, Wang, Zhang, Liu, Xie, and He}]{dgcnn}
\bibinfo{author}{Z.~Diao}, \bibinfo{author}{X.~Wang},
  \bibinfo{author}{D.~Zhang}, \bibinfo{author}{Y.~Liu},
  \bibinfo{author}{K.~Xie}, \bibinfo{author}{S.~He},
\newblock \bibinfo{title}{Dynamic spatial-temporal graph convolutional neural
  networks for traffic forecasting},
\newblock in: \bibinfo{booktitle}{Proceedings of the 33rd AAAI Conference on
  Artificial Intelligence}, \bibinfo{year}{2019}, pp.
  \bibinfo{pages}{890--897}. \DOIprefix\doi{10.1609/aaai.v33i01.3301890}.
\bibitem[{Cao et~al.(2020)Cao, Wang, Duan, Zhang, Zhu, Huang, Tong, Xu, Bai,
  Tong, and Zhang}]{stem}
\bibinfo{author}{D.~Cao}, \bibinfo{author}{Y.~Wang}, \bibinfo{author}{J.~Duan},
  \bibinfo{author}{C.~Zhang}, \bibinfo{author}{X.~Zhu},
  \bibinfo{author}{C.~Huang}, \bibinfo{author}{Y.~Tong},
  \bibinfo{author}{B.~Xu}, \bibinfo{author}{J.~Bai}, \bibinfo{author}{J.~Tong},
  \bibinfo{author}{Q.~Zhang},
\newblock \bibinfo{title}{Spectral temporal graph neural network for
  multivariate time-series forecasting},
\newblock in: \bibinfo{booktitle}{Proceedings of the 34th International
  Conference on Neural Information Processing Systems},
  volume~\bibinfo{volume}{33}, \bibinfo{year}{2020}, pp.
  \bibinfo{pages}{17766--17778}. \URLprefix
  \url{https://proceedings.neurips.cc/paper_files/paper/2020/file/cdf6581cb7aca4b7e19ef136c6e601a5-Paper.pdf}.
\bibitem[{Li and Zhu(2021)}]{sftgnn}
\bibinfo{author}{M.~Li}, \bibinfo{author}{Z.~Zhu},
\newblock \bibinfo{title}{Spatial-temporal fusion graph neural networks for
  traffic flow forecasting},
\newblock in: \bibinfo{booktitle}{Proceedings of the 35th AAAI Conference on
  Artificial Intelligence}, \bibinfo{year}{2021}, pp.
  \bibinfo{pages}{4189--4196}. \DOIprefix\doi{10.1609/aaai.v35i5.16542}.
\bibitem[{Chen et~al.(2021)Chen, Segovia, and Gel}]{zigzag}
\bibinfo{author}{Y.~Chen}, \bibinfo{author}{I.~Segovia}, \bibinfo{author}{Y.~R.
  Gel},
\newblock \bibinfo{title}{Z-gcnets: Time zigzags at graph convolutional
  networks for time series forecasting},
\newblock in: \bibinfo{booktitle}{Proceedings of the 38th International
  Conference on Machine Learning}, \bibinfo{year}{2021}, pp.
  \bibinfo{pages}{1684--1694}. \URLprefix
  \url{https://proceedings.mlr.press/v139/chen21o.html}.
\bibitem[{Guo et~al.(2021)Guo, Hu, Sun, Qian, Gao, and Yin}]{hgcn}
\bibinfo{author}{K.~Guo}, \bibinfo{author}{Y.~Hu}, \bibinfo{author}{Y.~Sun},
  \bibinfo{author}{S.~Qian}, \bibinfo{author}{J.~Gao},
  \bibinfo{author}{B.~Yin},
\newblock \bibinfo{title}{Hierarchical graph convolution network for traffic
  forecasting},
\newblock in: \bibinfo{booktitle}{Proceedings of the 35th AAAI Conference on
  Artificial Intelligence}, \bibinfo{year}{2021}, pp.
  \bibinfo{pages}{151--159}. \DOIprefix\doi{10.1609/aaai.v35i1.16088}.
\bibitem[{Lu et~al.(2020)Lu, Gan, Jin, Fu, and Zhang}]{stag}
\bibinfo{author}{B.~Lu}, \bibinfo{author}{X.~Gan}, \bibinfo{author}{H.~Jin},
  \bibinfo{author}{L.~Fu}, \bibinfo{author}{H.~Zhang},
\newblock \bibinfo{title}{Spatiotemporal adaptive gated graph convolution
  network for urban traffic flow forecasting},
\newblock in: \bibinfo{booktitle}{Proceedings of the 29th ACM International
  Conference on Information \& Knowledge Management}, \bibinfo{year}{2020}, pp.
  \bibinfo{pages}{1025--1034}. \DOIprefix\doi{10.1145/3340531.3411894}.
\bibitem[{Ye et~al.(2021)Ye, Sun, Du, Fu, and Xiong}]{cgc}
\bibinfo{author}{J.~Ye}, \bibinfo{author}{L.~Sun}, \bibinfo{author}{B.~Du},
  \bibinfo{author}{Y.~Fu}, \bibinfo{author}{H.~Xiong},
\newblock \bibinfo{title}{Coupled layer-wise graph convolution for
  transportation demand prediction},
\newblock \bibinfo{year}{2021}, pp. \bibinfo{pages}{4617--4625}.
  \DOIprefix\doi{10.1609/aaai.v35i5.16591}.
\bibitem[{Zhao et~al.(2020)Zhao, Song, Zhang, Liu, Wang, Lin, Deng, and
  Li}]{tgcn}
\bibinfo{author}{L.~Zhao}, \bibinfo{author}{Y.~Song},
  \bibinfo{author}{C.~Zhang}, \bibinfo{author}{Y.~Liu},
  \bibinfo{author}{P.~Wang}, \bibinfo{author}{T.~Lin},
  \bibinfo{author}{M.~Deng}, \bibinfo{author}{H.~Li},
\newblock \bibinfo{title}{T-gcn: A temporal graph convolutional network for
  traffic prediction},
\newblock \bibinfo{journal}{IEEE Transactions on Intelligent Transportation
  Systems} \bibinfo{volume}{21} (\bibinfo{year}{2020})
  \bibinfo{pages}{3848--3858}. \DOIprefix\doi{10.1109/TITS.2019.2935152}.
\bibitem[{Wu et~al.(2019)Wu, Pan, Long, Jiang, and Zhang}]{gwnet}
\bibinfo{author}{Z.~Wu}, \bibinfo{author}{S.~Pan}, \bibinfo{author}{G.~Long},
  \bibinfo{author}{J.~Jiang}, \bibinfo{author}{C.~Zhang},
\newblock \bibinfo{title}{Graph wavenet for deep spatial-temporal graph
  modeling},
\newblock in: \bibinfo{booktitle}{Proceedings of the 28th International Joint
  Conference on Artificial Intelligence}, \bibinfo{year}{2019}, p.
  \bibinfo{pages}{1907–1913}. \URLprefix
  \url{https://dl.acm.org/doi/abs/10.5555/3367243.3367303}.
\bibitem[{Zhou et~al.(2021)Zhou, Zhang, Peng, Zhang, Li, Xiong, and
  Zhang}]{info}
\bibinfo{author}{H.~Zhou}, \bibinfo{author}{S.~Zhang},
  \bibinfo{author}{J.~Peng}, \bibinfo{author}{S.~Zhang},
  \bibinfo{author}{J.~Li}, \bibinfo{author}{H.~Xiong},
  \bibinfo{author}{W.~Zhang},
\newblock \bibinfo{title}{Informer: Beyond efficient transformer for long
  sequence time-series forecasting},
\newblock in: \bibinfo{booktitle}{Proceedings of the 35th AAAI Conference on
  Artificial Intelligence}, \bibinfo{year}{2021}, pp.
  \bibinfo{pages}{11106--11115}. \DOIprefix\doi{10.1609/aaai.v35i12.17325}.
\bibitem[{Zhou et~al.(2022)Zhou, Ma, Wen, Wang, Sun, and Jin}]{fed}
\bibinfo{author}{T.~Zhou}, \bibinfo{author}{Z.~Ma}, \bibinfo{author}{Q.~Wen},
  \bibinfo{author}{X.~Wang}, \bibinfo{author}{L.~Sun},
  \bibinfo{author}{R.~Jin},
\newblock \bibinfo{title}{{FED}former: Frequency enhanced decomposed
  transformer for long-term series forecasting},
\newblock in: \bibinfo{booktitle}{Proceedings of the 39th International
  Conference on Machine Learning}, \bibinfo{year}{2022}, pp.
  \bibinfo{pages}{27268--27286}. \URLprefix
  \url{https://proceedings.mlr.press/v162/zhou22g.html}.
\bibitem[{Liu et~al.(2022)Liu, Yu, Liao, Li, Lin, Liu, and Dustdar}]{pyra}
\bibinfo{author}{S.~Liu}, \bibinfo{author}{H.~Yu}, \bibinfo{author}{C.~Liao},
  \bibinfo{author}{J.~Li}, \bibinfo{author}{W.~Lin}, \bibinfo{author}{A.~X.
  Liu}, \bibinfo{author}{S.~Dustdar},
\newblock \bibinfo{title}{Pyraformer: Low-complexity pyramidal attention for
  long-range time series modeling and forecasting},
\newblock in: \bibinfo{booktitle}{Proceedings of the 10th International
  Conference on Learning Representations}, \bibinfo{year}{2022}. \URLprefix
  \url{https://openreview.net/forum?id=0EXmFzUn5I}.
\bibitem[{Lai et~al.(2018)Lai, Chang, Yang, and Liu}]{lstnet}
\bibinfo{author}{G.~Lai}, \bibinfo{author}{W.-C. Chang},
  \bibinfo{author}{Y.~Yang}, \bibinfo{author}{H.~Liu},
\newblock \bibinfo{title}{Modeling long- and short-term temporal patterns with
  deep neural networks},
\newblock in: \bibinfo{booktitle}{Proceedings of the 42th International ACM
  SIGIR Conference on Research and Development in Information Retrieval},
  \bibinfo{year}{2018}, p. \bibinfo{pages}{95–104}.
  \DOIprefix\doi{10.1145/3209978.3210006}.
\bibitem[{Vaswani et~al.(2017)Vaswani, Shazeer, Parmar, Uszkoreit, Jones,
  Gomez, Kaiser, and Polosukhin}]{att}
\bibinfo{author}{A.~Vaswani}, \bibinfo{author}{N.~Shazeer},
  \bibinfo{author}{N.~Parmar}, \bibinfo{author}{J.~Uszkoreit},
  \bibinfo{author}{L.~Jones}, \bibinfo{author}{A.~N. Gomez},
  \bibinfo{author}{{\L}.~Kaiser}, \bibinfo{author}{I.~Polosukhin},
\newblock \bibinfo{title}{Attention is all you need},
\newblock in: \bibinfo{booktitle}{Proceedings of the 31st International
  Conference on Neural Information Processing Systems}, \bibinfo{year}{2017},
  pp. \bibinfo{pages}{6000--6010}.
\bibitem[{Guo et~al.(2019)Guo, Lin, Feng, Song, and Wan}]{astgcn}
\bibinfo{author}{S.~Guo}, \bibinfo{author}{Y.~Lin}, \bibinfo{author}{N.~Feng},
  \bibinfo{author}{C.~Song}, \bibinfo{author}{H.~Wan},
\newblock \bibinfo{title}{Attention based spatial-temporal graph convolutional
  networks for traffic flow forecasting},
\newblock in: \bibinfo{booktitle}{Proceedings of the 33rd AAAI Conference on
  Artificial Intelligence}, \bibinfo{year}{2019}, pp.
  \bibinfo{pages}{922--929}. \DOIprefix\doi{10.1609/aaai.v33i01.3301922}.
\bibitem[{Cui et~al.(2021)Cui, Xie, and Zheng}]{hi}
\bibinfo{author}{Y.~Cui}, \bibinfo{author}{J.~Xie}, \bibinfo{author}{K.~Zheng},
\newblock \bibinfo{title}{Historical inertia: A neglected but powerful baseline
  for long sequence time-series forecasting},
\newblock in: \bibinfo{booktitle}{Proceedings of the 30th ACM International
  Conference on Information \& Knowledge Management}, \bibinfo{year}{2021}, p.
  \bibinfo{pages}{2965–2969}. \DOIprefix\doi{10.1145/3459637.3482120}.
\bibitem[{Cirstea et~al.(2022)Cirstea, Yang, Guo, Kieu, and Pan}]{agnostic}
\bibinfo{author}{R.-G. Cirstea}, \bibinfo{author}{B.~Yang},
  \bibinfo{author}{C.~Guo}, \bibinfo{author}{T.~Kieu},
  \bibinfo{author}{S.~Pan},
\newblock \bibinfo{title}{Towards spatio- temporal aware traffic time series
  forecasting},
\newblock in: \bibinfo{booktitle}{Proceedings of the IEEE 38th International
  Conference on Data Engineering}, \bibinfo{year}{2022}, pp.
  \bibinfo{pages}{2900--2913}. \DOIprefix\doi{10.1109/ICDE53745.2022.00262}.
\bibitem[{Gu et~al.(2022)Gu, Goel, and Re}]{s4}
\bibinfo{author}{A.~Gu}, \bibinfo{author}{K.~Goel}, \bibinfo{author}{C.~Re},
\newblock \bibinfo{title}{Efficiently modeling long sequences with structured
  state spaces},
\newblock \bibinfo{journal}{arXiv preprint arXiv:2111.00396}
  (\bibinfo{year}{2022}). \DOIprefix\doi{10.48550/arXiv.2111.00396}.
\bibitem[{Gu and Dao(2023)}]{mamba}
\bibinfo{author}{A.~Gu}, \bibinfo{author}{T.~Dao},
\newblock \bibinfo{title}{Mamba: Linear-time sequence modeling with selective
  state spaces},
\newblock \bibinfo{journal}{arXiv preprint arXiv:2312.00752}
  (\bibinfo{year}{2023}). \DOIprefix\doi{10.48550/arXiv.2312.00752}.
\bibitem[{Wang et~al.(2024)Wang, Tsepa, Ma, and Wang}]{gmamba1}
\bibinfo{author}{C.~Wang}, \bibinfo{author}{O.~Tsepa}, \bibinfo{author}{J.~Ma},
  \bibinfo{author}{B.~Wang},
\newblock \bibinfo{title}{Graph-mamba: Towards long-range graph sequence
  modeling with selective state spaces},
\newblock \bibinfo{journal}{arXiv preprint arXiv:2402.00789}
  (\bibinfo{year}{2024}). \DOIprefix\doi{10.48550/arXiv.2402.00789}.
\bibitem[{Behrouz and Hashemi(2024)}]{gmamba2}
\bibinfo{author}{A.~Behrouz}, \bibinfo{author}{F.~Hashemi},
\newblock \bibinfo{title}{Graph mamba: Towards learning on graphs with state
  space models},
\newblock \bibinfo{journal}{arXiv preprint arXiv:2402.08678}
  (\bibinfo{year}{2024}). \DOIprefix\doi{10.48550/arXiv.2402.08678}.
\bibitem[{Li et~al.(2024)Li, Wang, Zhang, and Coster}]{stgmamba}
\bibinfo{author}{L.~Li}, \bibinfo{author}{H.~Wang}, \bibinfo{author}{W.~Zhang},
  \bibinfo{author}{A.~Coster},
\newblock \bibinfo{title}{Stg-mamba: Spatial-temporal graph learning via
  selective state space model},
\newblock \bibinfo{journal}{arXiv preprint arXiv:2403.12418}
  (\bibinfo{year}{2024}). \DOIprefix\doi{10.48550/arXiv.2403.12418}.
\bibitem[{Fang et~al.(2021)Fang, Long, Song, and Xie}]{stgode}
\bibinfo{author}{Z.~Fang}, \bibinfo{author}{Q.~Long},
  \bibinfo{author}{G.~Song}, \bibinfo{author}{K.~Xie},
\newblock \bibinfo{title}{Spatial-temporal graph ode networks for traffic flow
  forecasting},
\newblock in: \bibinfo{booktitle}{Proceedings of the 27th ACM SIGKDD Conference
  on Knowledge Discovery \& Data Mining}, \bibinfo{year}{2021}, pp.
  \bibinfo{pages}{364–--373}. \DOIprefix\doi{10.1145/3447548.3467430}.
\bibitem[{Liu et~al.(2023)Liu, Dong, Jiang, Deng, Deng, Chen, and Song}]{stae}
\bibinfo{author}{H.~Liu}, \bibinfo{author}{Z.~Dong},
  \bibinfo{author}{R.~Jiang}, \bibinfo{author}{J.~Deng},
  \bibinfo{author}{J.~Deng}, \bibinfo{author}{Q.~Chen},
  \bibinfo{author}{X.~Song},
\newblock \bibinfo{title}{Spatio-temporal adaptive embedding makes vanilla
  transformer sota for traffic forecasting},
\newblock in: \bibinfo{booktitle}{Proceedings of the 32nd ACM International
  Conference on Information and Knowledge Management}, \bibinfo{year}{2023}, p.
  \bibinfo{pages}{4125–4129}. \DOIprefix\doi{10.1145/3583780.3615160}.
\bibitem[{Song et~al.(2020)Song, Lin, Guo, and Wan}]{stsgcn}
\bibinfo{author}{C.~Song}, \bibinfo{author}{Y.~Lin}, \bibinfo{author}{S.~Guo},
  \bibinfo{author}{H.~Wan},
\newblock \bibinfo{title}{Spatial-temporal synchronous graph convolutional
  networks: A new framework for spatial-temporal network data forecasting},
\newblock in: \bibinfo{booktitle}{Proceedings of the 34th AAAI Conference on
  Artificial Intelligence}, \bibinfo{year}{2020}, pp.
  \bibinfo{pages}{914--921}. \DOIprefix\doi{10.1609/aaai.v34i01.5438}.
\bibitem[{Li et~al.(2018)Li, Yu, Shahabi, and Liu}]{dcrnn}
\bibinfo{author}{Y.~Li}, \bibinfo{author}{R.~Yu}, \bibinfo{author}{C.~Shahabi},
  \bibinfo{author}{Y.~Liu},
\newblock \bibinfo{title}{Diffusion convolutional recurrent neural network:
  Data-driven traffic forecasting},
\newblock in: \bibinfo{booktitle}{Proceedings of the 6th International
  Conference on Learning Representations}, \bibinfo{year}{2018}. \URLprefix
  \url{https://openreview.net/forum?id=SJiHXGWAZ}.
\bibitem[{Bai et~al.(2020)Bai, Yao, Li, Wang, and Wang}]{agcrn}
\bibinfo{author}{L.~Bai}, \bibinfo{author}{L.~Yao}, \bibinfo{author}{C.~Li},
  \bibinfo{author}{X.~Wang}, \bibinfo{author}{C.~Wang},
\newblock \bibinfo{title}{Adaptive graph convolutional recurrent network for
  traffic forecasting},
\newblock in: \bibinfo{booktitle}{Proceedings of the 34th International
  Conference on Neural Information Processing Systems}, \bibinfo{year}{2020},
  p. \bibinfo{pages}{17804–17815}. \URLprefix
  \url{https://dl.acm.org/doi/abs/10.5555/3495724.3497218}.
\bibitem[{Shang et~al.(2021)Shang, Chen, and Bi}]{gts}
\bibinfo{author}{C.~Shang}, \bibinfo{author}{J.~Chen}, \bibinfo{author}{J.~Bi},
\newblock \bibinfo{title}{Discrete graph structure learning for forecasting
  multiple time series},
\newblock in: \bibinfo{booktitle}{Proceedings of the 9th International
  Conference on Learning Representations}, \bibinfo{year}{2021}. \URLprefix
  \url{https://openreview.net/forum?id=WEHSlH5mOk}.
\bibitem[{Wu et~al.(2020)Wu, Pan, Long, Jiang, Chang, and Zhang}]{mtgnn}
\bibinfo{author}{Z.~Wu}, \bibinfo{author}{S.~Pan}, \bibinfo{author}{G.~Long},
  \bibinfo{author}{J.~Jiang}, \bibinfo{author}{X.~Chang},
  \bibinfo{author}{C.~Zhang},
\newblock \bibinfo{title}{Connecting the dots: Multivariate time series
  forecasting with graph neural networks},
\newblock in: \bibinfo{booktitle}{Proceedings of the 26th ACM SIGKDD
  International Conference on Knowledge Discovery \& Data Mining},
  \bibinfo{year}{2020}, p. \bibinfo{pages}{753–763}.
  \DOIprefix\doi{10.1145/3394486.3403118}.
\bibitem[{Zheng et~al.(2020)Zheng, Fan, Wang, and Qi}]{gman}
\bibinfo{author}{C.~Zheng}, \bibinfo{author}{X.~Fan},
  \bibinfo{author}{C.~Wang}, \bibinfo{author}{J.~Qi},
\newblock \bibinfo{title}{Gman: A graph multi-attention network for traffic
  prediction},
\newblock in: \bibinfo{booktitle}{Proceedings of the 34th AAAI Conference on
  Artificial Intelligence}, \bibinfo{year}{2020}, pp.
  \bibinfo{pages}{1234--1241}. \DOIprefix\doi{10.1609/aaai.v34i01.5477}.
\bibitem[{Jiang et~al.(2023)Jiang, Han, Zhao, and Wang}]{pd}
\bibinfo{author}{J.~Jiang}, \bibinfo{author}{C.~Han}, \bibinfo{author}{W.~X.
  Zhao}, \bibinfo{author}{J.~Wang},
\newblock \bibinfo{title}{Pdformer: Propagation delay-aware dynamic long-range
  transformer for traffic flow prediction},
\newblock in: \bibinfo{booktitle}{Proceedings of the 37th AAAI Conference on
  Artificial Intelligence}, \bibinfo{year}{2023}, pp.
  \bibinfo{pages}{4365--4373}. \DOIprefix\doi{10.1609/aaai.v37i4.25556}.
\bibitem[{Deng et~al.(2021)Deng, Chen, Jiang, Song, and Tsang}]{stnorm}
\bibinfo{author}{J.~Deng}, \bibinfo{author}{X.~Chen},
  \bibinfo{author}{R.~Jiang}, \bibinfo{author}{X.~Song}, \bibinfo{author}{I.~W.
  Tsang},
\newblock \bibinfo{title}{St-norm: Spatial and temporal normalization for
  multi-variate time series forecasting},
\newblock in: \bibinfo{booktitle}{Proceedings of the 27th ACM SIGKDD Conference
  on Knowledge Discovery \& Data Mining}, \bibinfo{year}{2021}, p.
  \bibinfo{pages}{269–278}. \DOIprefix\doi{10.1145/3447548.3467330}.
\bibitem[{Shao et~al.(2022)Shao, Zhang, Wang, Wei, and Xu}]{stid}
\bibinfo{author}{Z.~Shao}, \bibinfo{author}{Z.~Zhang},
  \bibinfo{author}{F.~Wang}, \bibinfo{author}{W.~Wei}, \bibinfo{author}{Y.~Xu},
\newblock \bibinfo{title}{Spatial-temporal identity: A simple yet effective
  baseline for multivariate time series forecasting},
\newblock in: \bibinfo{booktitle}{Proceedings of the 31st ACM International
  Conference on Information \& Knowledge Management}, \bibinfo{year}{2022}, p.
  \bibinfo{pages}{4454–4458}. \DOIprefix\doi{10.1145/3511808.3557702}.
\bibitem[{Wang et~al.(2020)Wang, Zheng, Ye, Gan, Li, Song, Zhou, Ma, Yu, Gai,
  Xiao, He, Karypis, Li, and Zhang}]{dgl}
\bibinfo{author}{M.~Wang}, \bibinfo{author}{D.~Zheng}, \bibinfo{author}{Z.~Ye},
  \bibinfo{author}{Q.~Gan}, \bibinfo{author}{M.~Li}, \bibinfo{author}{X.~Song},
  \bibinfo{author}{J.~Zhou}, \bibinfo{author}{C.~Ma}, \bibinfo{author}{L.~Yu},
  \bibinfo{author}{Y.~Gai}, \bibinfo{author}{T.~Xiao}, \bibinfo{author}{T.~He},
  \bibinfo{author}{G.~Karypis}, \bibinfo{author}{J.~Li},
  \bibinfo{author}{Z.~Zhang},
\newblock \bibinfo{title}{Deep graph library: A graph-centric,
  highly-performant package for graph neural networks},
\newblock \bibinfo{journal}{arXiv preprint arXiv:1909.01315}
  (\bibinfo{year}{2020}). \DOIprefix\doi{10.48550/arXiv.1909.01315}.
\bibitem[{Paszke et~al.(2019)Paszke, Gross, Massa, Lerer, Bradbury, Chanan,
  Killeen, Lin, Gimelshein, Antiga, Desmaison, Kopf, Yang, DeVito, Raison,
  Tejani, Chilamkurthy, Steiner, Fang, Bai, and Chintala}]{pytorch}
\bibinfo{author}{A.~Paszke}, \bibinfo{author}{S.~Gross},
  \bibinfo{author}{F.~Massa}, \bibinfo{author}{A.~Lerer},
  \bibinfo{author}{J.~Bradbury}, \bibinfo{author}{G.~Chanan},
  \bibinfo{author}{T.~Killeen}, \bibinfo{author}{Z.~Lin},
  \bibinfo{author}{N.~Gimelshein}, \bibinfo{author}{L.~Antiga},
  \bibinfo{author}{A.~Desmaison}, \bibinfo{author}{A.~Kopf},
  \bibinfo{author}{E.~Yang}, \bibinfo{author}{Z.~DeVito},
  \bibinfo{author}{M.~Raison}, \bibinfo{author}{A.~Tejani},
  \bibinfo{author}{S.~Chilamkurthy}, \bibinfo{author}{B.~Steiner},
  \bibinfo{author}{L.~Fang}, \bibinfo{author}{J.~Bai},
  \bibinfo{author}{S.~Chintala},
\newblock \bibinfo{title}{Pytorch: An imperative style, high-performance deep
  learning library},
\newblock in: \bibinfo{booktitle}{Proceedings of the 33th International
  Conference on Neural Information Processing Systems}, \bibinfo{year}{2019},
  pp. \bibinfo{pages}{8024--8035}. \URLprefix
  \url{https://proceedings.neurips.cc/paper_files/paper/2019/file/bdbca288fee7f92f2bfa9f7012727740-Paper.pdf}.
\bibitem[{Kingma and Ba(2015)}]{adam}
\bibinfo{author}{D.~P. Kingma}, \bibinfo{author}{J.~Ba},
\newblock \bibinfo{title}{Adam: A method for stochastic optimization},
\newblock in: \bibinfo{booktitle}{Proceedings of the 3rd International
  Conference on Learning Representations}, \bibinfo{year}{2015}. \URLprefix
  \url{https://doi.org/10.48550/arXiv.1412.6980}.
\bibitem[{Kim et~al.(2023)Kim, Choi, and Whang}]{drag}
\bibinfo{author}{H.~Kim}, \bibinfo{author}{J.~Choi}, \bibinfo{author}{J.~J.
  Whang},
\newblock \bibinfo{title}{Dynamic relation-attentive graph neural networks for
  fraud detection},
\newblock in: \bibinfo{booktitle}{Proceedings of 2023 IEEE International
  Conference on Data Mining Workshops (ICDMW)}, \bibinfo{year}{2023}, pp.
  \bibinfo{pages}{1092--1096}. \DOIprefix\doi{10.1109/ICDMW60847.2023.00143}.
\bibitem[{Chung et~al.(2023)Chung, Lee, and Whang}]{hynt}
\bibinfo{author}{C.~Chung}, \bibinfo{author}{J.~Lee}, \bibinfo{author}{J.~J.
  Whang},
\newblock \bibinfo{title}{Representation learning on hyper-relational and
  numeric knowledge graphs with transformers},
\newblock in: \bibinfo{booktitle}{Proceedings of the 29th ACM SIGKDD Conference
  on Knowledge Discovery and Data Mining}, \bibinfo{year}{2023}, pp.
  \bibinfo{pages}{310--322}.
\bibitem[{Huang et~al.(2022)Huang, Yang, Wang, Wang, Zhang, Xu, Chen, and
  Vazirgiannis}]{dgraph}
\bibinfo{author}{X.~Huang}, \bibinfo{author}{Y.~Yang},
  \bibinfo{author}{Y.~Wang}, \bibinfo{author}{C.~Wang},
  \bibinfo{author}{Z.~Zhang}, \bibinfo{author}{J.~Xu},
  \bibinfo{author}{L.~Chen}, \bibinfo{author}{M.~Vazirgiannis},
\newblock \bibinfo{title}{Dgraph: A large-scale financial dataset for graph
  anomaly detection},
\newblock in: \bibinfo{booktitle}{Proceedings of the 36th International
  Conference on Neural Information Processing Systems}, \bibinfo{year}{2022},
  pp. \bibinfo{pages}{22765--22777}. \URLprefix
  \url{https://proceedings.neurips.cc/paper_files/paper/2022/file/8f1918f71972789db39ec0d85bb31110-Paper-Datasets_and_Benchmarks.pdf}.
\bibitem[{Lee et~al.(2023)Lee, Chung, and Whang}]{ingram}
\bibinfo{author}{J.~Lee}, \bibinfo{author}{C.~Chung}, \bibinfo{author}{J.~J.
  Whang},
\newblock \bibinfo{title}{{InGram}: Inductive knowledge graph embedding via
  relation graphs},
\newblock in: \bibinfo{booktitle}{Proceedings of the 40th International
  Conference on Machine Learning}, \bibinfo{year}{2023}, pp.
  \bibinfo{pages}{18796--18809}.

\end{thebibliography}

\end{document}